\documentclass[sigconf]{acmart}

\usepackage{multirow} 

\AtBeginDocument{%
  }

\setcopyright{acmlicensed}
\copyrightyear{2025}
\acmYear{2025}
\acmConference[ACM ICAIF 2025]{AI for Finance Symposium '25, The 2nd Workshop on LLMs and Generative AI for Finance}{November 15,
  2025}{Singapore}

\begin{document}

\title{Open Banking Foundational Model: Learning Language Representations from Few Financial Transactions}

\author{Gustavo Polleti}
\email{gustavo.polleti@trustly.com}
\orcid{1234-5678-9012}
\affiliation{%
  \institution{Trustly}
  \city{S\~ao Paulo}
  \state{S\~ao Paulo}
  \country{Brazil}
}

\author{Marlesson Santana}
\email{marlesson.santana@trustly.com}
\orcid{1234-5678-9012}
\affiliation{%
  \institution{Trustly}
  \city{Ilhéus}
  \state{Bahia}
  \country{Brazil}
}

\author{Eduardo Fontes}
\email{eduardo.fontes@trustly.com}
\orcid{1234-5678-9012}
\affiliation{%
  \institution{Trustly}
  \city{Concord}
  \state{North Carolina}
  \country{USA}
}

\renewcommand{\shortauthors}{Polleti et al.}

\begin{abstract}
  We introduced a multimodal foundational model for financial transactions that integrates both structured attributes and unstructured textual descriptions into a unified representation. By adapting masked language modeling to transaction sequences, we demonstrated that our approach not only outperforms classical feature engineering and discrete event sequence methods but is also particularly effective in data-scarce Open Banking scenarios. To our knowledge, this is the first large-scale study across thousands of financial institutions in North America, providing evidence that multimodal representations can generalize across geographies and institutions. These results highlight the potential of self-supervised models to advance financial applications ranging from fraud prevention and credit risk to customer insights.
\end{abstract}





\maketitle

\section{Introduction}

A lot can be learned from how someone spends their money. A customer self reported address can be challenged if they recently purchased at stores in a different location~\cite{transgeo23}. A financial institution can extract meaningful insights about their customers by looking at their financial transactions activity history. Not surprisingly, transaction history has been the backbone for several applications in the financial domain. Fraud detection~\cite{JHA201212650}, money laundering detection~\cite{Dumitry22}, credit default prediction, customer churn prediction~\cite{NPPR_24}, and future expenditure modelling~\cite{transgeo23} are classical examples. Financial transactions are often modeled as discrete event sequences. Each transaction is often represented as a single event, and a customer account as an event timeseries. Many methods have been proposed to represent these event timeseries as embeddings that are useful to support typical financial applications~\cite{coles_22}. Embeddings appeal lies on being able to replace the usually expensive and time-consuming feature engineering process in favour of learnt representations. Such methods often take the financial transactions history as a multivariate timeseries, where each transaction contains a set of numerical or categorical values. For example, a single transaction can be described with tabular features, such as its timestamp, amount, card number, merchant name or category. The customer account is described as their transactions history, which becomes an ordered transactions set. Then, these methods often train a self-supervised foundational model to produce embeddings that are later used as input for downstream tasks, e.g. a binary classifier to tell if a customer account is fraudulent or not. Such methods, despite outperforming or being on par with classical machine learning methods based on hand-crafted features~\cite{NPPR_24, coles_22}, they are restricted to multivariate event representation and, thus, miss on fully benefiting from the unstructured data that often escort transactions. Financial transactions may be escorted by natural language descriptions that do not strictly follow proper standards or may vary depending on the geography, financial institution or context~\cite{catcoin20}. Unlike established formats, like MCC codes for credit card transactions or merchant codes for registered institutions, bank account transactions contain unstructured information that may be hard to represent as numerical or categorical features. If we require to feature engineer transactions textual descriptions into numerical or categorical features beforehand, embeddings lose part of their appeal. Furthermore, benchmarks that consider transaction multimodality are rare. Multimodal Banking Dataset (MDB) has recently been released but still lacks raw textual descriptions to support research on this front~\cite{mdb25}. Benchmarks that include transaction textual descriptions often face privacy concerns due to the sensitive nature of financial data. Since textual descriptions are unstructured, even with advanced anonymization techniques, it is hard to guarantee no personally identifiable information (PII) is leaked. However, even if such a benchmark could be produced, since financial transactions textual descriptions may vary a lot depending on the financial institution or geography, it may arise questions around if the benchmark generalizes across different insitutions.

Open Banking represents a great opportunity to produce both benchmarks and foundational models that generalize across different financial institutions and geographies. In this work, we introduce a multimodal foundational model trained on financial transaction sequences from 10,000 financial institutions in North America, both United States and Canada. We built a private dataset featuring financial transaction records of approximately 10 million accounts, routing code and account number unique pairs (ARN). Each ARN contains up to 90 days transactions history. Each transaction contains a textual description, timestamp and amount value. We adapted promising techniques that learns language representations for sequencial recommendation~\cite{textisall23, leveragingllm23} to our domain inspired by early success in the financial industry~\footnote{https://building.nubank.com/defining-an-interface-between-transaction-data-and-foundation-models/}. We propose to represent a transaction as a ``sentence'', combining both tabular attributes and the description as a single text, so that a transaction history for an ARN becomes a document, a sequence of sentences. Our foundational model is then finetuned on our private dataset using bi-directional attention, based on BERT models~\cite{BERT17}, with masked language modeling. In this setup, the pretraining is oriented to infer masked tokens based on the ARN financial transactions context in a self-supervised fashion. To assert usefulness of our learnt representations, we designed 19 downstream tasks, ranging from typical financial applications (e.g. credit risk and fraud prevention) and customer demographics (gender, age, location) to ``banking'' tasks. Our banking tasks are related to assess whether our embeddings capture bank account characteristics that are agnostic to financial institutions, for example, account type (checkings or savings), account profile (business or personal), or even if we can infer which financial institution the ARN is associated with. As far as we could find, this is the first large-scale study of financial bank activity data across multiple financial institutions in North America. Finally, we present empirical evidence that our methods can outperform both classical machine learning models, which rely on handcrafted features, and other discrete event sequence embedding methods in scenarios where only a few financial transactions are available. 

In particular, we argue that our methods are most notable in data scarcity scenarios where we don't have many financial transaction events available for a given account. In North America, you typically have to perform multiple Open Baking API calls to retrieve the transactions history for a given account, which can increase latency and cost. As a result, you often have just a few financial transctions available. Traditional methods that rely solely on discrete event sequences may require more events than are available to perform well in Open Banking scenarios where data is scarce.




\section{Our Approach}

We propose representing a financial transaction as a ``sentence'' composed of structured events in the format "\texttt{[TYPE] <DEBIT|CREDIT> [AMT] <AMOUNT> [NAME] <DESCRIPTION>}", combining both tabular attributes and the transaction description into a single textual sequence. Roughly, we are representing account transactions history as documents in a specific language and, then, training language models. Our foundational language model is fine-tuned on a private dataset using bi-directional attention, following the BERT-based architecture~\cite{BERT17}, with masked language modeling (MLM) trained to predict masked tokens (i.e., \texttt{<DEBIT|CREDIT>}, \texttt{<AMOUNT>}, \texttt{<DESCRIPTION>}) from the surrounding context of financial transactions in a self-supervised manner.

Let a bank account be denoted as a sequence of transactions $X_{\text{raw}} = (x_1, x_2, x_3, \ldots, x_n)$, where each $x_k$ represents an individual transaction comprising the amount, description, and type fields. We formalize the representation of each transaction $x_k$ as a \textit{sentence} $s_k$ in a structured text format:
\[
s_k = \texttt{[TYPE]} \; t_k \; \texttt{[AMT]} \; a_k \; \texttt{[NAME]} \; d_k,
\]
where $t_k \in \{\texttt{DEBIT}, \texttt{CREDIT}\}$ denotes the transaction direction, $a_k \in \mathcal{A}$ is the transaction amount discretized into predefined buckets (e.g., \$50 intervals), and $d_k \in \mathcal{D}$ is the raw transaction description tokenized into subword units.

An ARN \textit{account history} is then modeled as a \textit{document}, i.e., a sequence of such sentences concatenated with a special separator token:
\[
D_i = (s_1 \; \texttt{[SEP]} \; s_2 \; \texttt{[SEP]} \; \cdots \; \texttt{[SEP]} \; s_{n_i}),
\]
where $D_i$ corresponds to the $i$-th account and $n_i$ is the number of transactions in its history. This textualized representation preserves both the sequential ordering of events and their structured attributes while enabling the use of Transformer-based language models.

Finally, the \textit{training dataset} is expressed as a \textit{corpus} $\mathcal{C}$ of such documents:
\[
\mathcal{C} = \{D_1, D_2, \ldots, D_m\},
\]
where $m$ is the total number of accounts. Each document in $\mathcal{C}$ constitutes the transaction history of a distinct account, and the full corpus provides a large-scale, self-supervised resource for pretraining a foundational financial language model.

 Given a pretrained foundational language model, the representation of the entire account is derived from the contextual embedding of the \texttt{[CLS]} token, which serves as a global summary of the transaction sequence. This embedding is used as the fixed-length vector representation of the bank account.

\section{Experiments}


\subsection{Dataset}

We conducted our experiments on a private transaction data from 99\% financial institutions in US and Canada from a well-known Online Banking Company.

Our pretraining dataset comprises financial transaction and account activity histories of 10,000,000 unique bank accounts, ARNs, from January 1, 2024 to February 1, 2025. The dataset is anonymized to remove any personal identifiable data. Each bank activity record includes the transaction date, amount (positive for inflows, negative for outflows), and description. The corpus has 491,796,263 tokens. Despite large, typically, the documents consists in less than $120$ transactions. Only 2\% accounts have more than $700$ transactions.

The dataset used for downstream task evaluation comprises 385,012 unique bank accounts, ARNs, from February 1, 2025 to March 30, 2025.

\subsection{Evaluation Tasks}

We designed $19$ typical downstream tasks for the Open Banking domain. We divide our tasks in four groups: (1) Demographics, (2) Risk, (3) Banking and (4) Geolocation.

In the Demographics group, the ``gender'' task is a binary classification problem that predicts whether the account holder is male or female. The target labels were derived from the account owner’s first name, using the U.S. Social Security Administration (SSA) Baby Names Dataset\footnote{\url{https://www.ssa.gov/oact/babynames/}}. To ensure label quality, we included only accounts whose first names were unambiguously associated with a single gender in the SSA records. The ``1st name'' task is a multiclass classification over the 50 most frequent first names observed in our dataset. This task is designed to evaluate the model’s ability to capture signals associated with personal identity. The ``age'' task involves predicting the account holder’s age group, discretized into 9 buckets spanning from 18 to 100 years old. Each bucket covers a 10-year range (e.g., 18–19, 20–29, 30–39, etc.). The target labels for this task were derived from a sample of users who self-reported their date of birth.

The Risk group includes six binary classification tasks centered on ACH (Automated Clearing House) return codes, which serve as early indicators of credit and fraud risk. ACH is the backbone of electronic payments in the U.S., supporting both direct deposits (credits) and bill payments or transfers (debits). Unlike real-time payment systems, ACH transactions are processed in batches, typically settling over 1 to 3 business days. This delay introduces a risk window during which a customer may withdraw or move funds, leading to failed payments.

Each task in this group aims to predict whether any transaction from a given account will experience a specific type of ACH return code in the month following the observation window. A return code indicates that an attempted debit from the account was unsuccessful, often due to account status or user behavior. Non-Sufficient Funds ``\textit{nsf}'' is the most common return code in both the U.S. and Canada. It occurs when an ACH debit cannot be completed because the account lacks sufficient funds at the time of settlement. ``\textit{stop}'' captures whether the account holder has issued a stop payment order during the processing window, instructing the bank to reject a scheduled debit. ``\textit{unauth}'' reflects unauthorized transactions, where the account holder requests a chargeback after a debit has already been processed—often associated with fraud or disputed payments. ``\textit{frozen}'' indicates that the attempted debit targeted a frozen account, either due to actions taken by the financial institution (e.g., suspected fraud) or external interventions (e.g., legal restrictions). ``\textit{suf}'' (Sufficient Funds) is a derived task that predicts whether the account avoids any of the less common but critical return codes: stop, unauth, or frozen. ``\textit{ret}'' (Any Return) generalizes the prediction to include any return code, combining nsf, stop, unauth, and frozen. These tasks are inherently unbalanced, as return events are relatively rare but carry significant operational and financial consequences. To address this imbalance during training, we apply a straightforward undersampling strategy, selecting a 1:4 ratio of positive (returned) to negative (non-returned) examples to maintain signal while reducing class skew.

The Banking group evaluates account-level and institutional properties. The ``debit card'' task is a binary classification that predicts whether the user has an active debit card. The ``inc. (income)'' task predicts the user’s income bucket, discretized into 50 quantiles. Similarly, ``bal. (balance)'' estimates the average balance of the account, also bucketed into 50 quantiles. The ``fi'' task predicts the financial institution associated with the account, limited to the 50 most common ones. ``act type'' determines whether the account is a checking or savings account, while ``act prof. (profile)'' identifies whether the account is personal or business-related.

Finally, the Geolocation group focuses on predicting the user's location. Each task, ``state 1'', ``city 1'', ``state 2'', and ``city 2'', predicts the US state, CA province or city of the user from one of two geolocation providers. These tasks are formulated as multiclass classification problems over the 50 most common states or cities, aiming to test the model’s ability to infer location-related attributes from financial transactions. The dual provider structure allows for cross-validation of geolocation signals across different data sources.



\subsection{Compared Methods}

\begin{figure*}
    \centering
    \includegraphics[width=\linewidth, height=0.12\textheight]{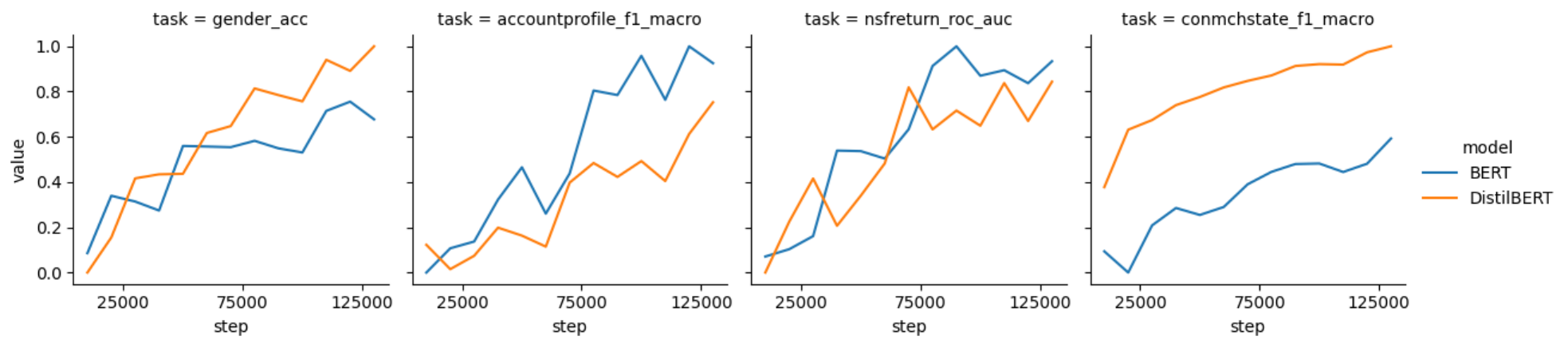}
    \caption{The downstream task performance during pretraining in normalized scale for two variants (BERT and DistilBERT). The (task, metric) pairs are: (gender, accuracy), (nsf, roc auc), (act prof., f1 macro), (state 1, f1 macro).}
    \label{fig:ssl_learning_curve}
\end{figure*}

We compare the following four methods: (1) \textbf{FeatEng}, (2) \textbf{CoLES}, (3) \textbf{BERT} and (4) \textbf{DistilBERT}. FeatEng uses the set of handcrafted features. For numerical features, we compute standard aggregation statistics, e.g. sum, count, mean, minimum, maximum, and standard deviation, across all transactions in an entity’s history. For categorical features, we group transactions by each unique value of the categorical variable and then compute the same numerical aggregations within each group. CoLES uses a contrastive learning approach in which positive pairs consist of random subsequences sampled from the same sequence, while negative pairs are formed by sampling subsequences from different sequences. The effectiveness of this method depends on the choice of minimum and maximum subsequence lengths used during sampling. To ensure consistency and comparability, we adopt the same values for these hyperparameters as specified in the original paper~\cite{coles_22}. BERT uses a masked language modeling (MLM) objective following the original BERT formulation, where random tokens in the input sequence are masked and the model is trained to predict the masked tokens based on their context. This approach enables the model to learn deep bidirectional representations of sequential data. We follow the standard pretraining setup described in the original BERT paper~\cite{BERT17} including masking strategy and model architecture. For the downstream task evaluation, we considered the ``CLS'' token embedding. Finally, we use the DistilBERT framework~\cite{DistilBERT19}, a lighter and faster variant of BERT trained using knowledge distillation. We use the same setup as for the BERT model.

Note, both BERT and DistilBERT have a 512 maximum context lenght, which limits how many transactions they will consider. Typically, the context window is able to consider less than 20 transactions, which can be considered data-scarce.

For each method, we applied a standard scaler on its features or embeddings, and then trained a logistic regression model for each downstream task. The linear model performance on each task is what we report in the next Section.

\section{Results}

Below, we present the results for each of the evaluated methods across all downstream tasks. Given our performance metrics may include proprietary information or expose competitive model behavior, we report only normalized scores, computed using min-max scaling across methods for each task. This transformation maps each metric to a [0, 1] range, where 0 represents the lowest and 1 the highest performance among all methods for a given task. Normalization allows for a fair and interpretable comparison between models while preserving the relative ranking of their performance.

\subsection{Pretraining Evaluation}

Figure~\ref{fig:ssl_learning_curve} displays the downstream task performance during the pretraining in normalized scale for 4 out of the 19 tasks, one for each group. Every 10k steps, for each model variant, we perform linear probing and record the performance. Despite not being monotonic, we observe the performance is consistently increasing as the pretraining continues. We observe the same for all tasks.

\subsection{Downstream Task Evaluation}

\begin{figure}
    \centering
    \includegraphics[width=\linewidth, height=0.12\textheight]{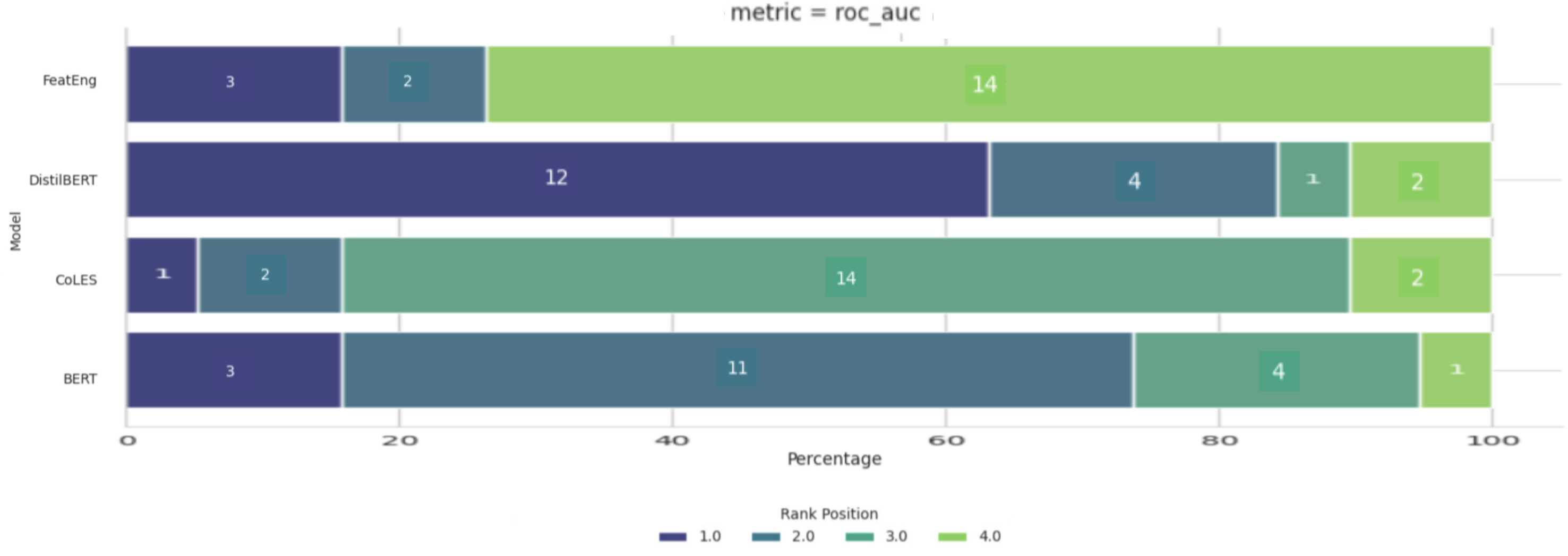}
    \caption{The downstream task performance rank distribution across all metrics.}
    \label{fig:rank_distribution_summary}
\end{figure}

Figure~\ref{fig:rank_distribution_summary} displays the distribution of rank positions according to all metrics for each method on the 19 downstream tasks. We observe that the DistilBERT outperforms other methods in most tasks across all metrics with BERT model catching up. Both often outperform CoLES and the feature engineering based method. These ranks were computed considering the embeddings after the pretraining without any additional finetuning.

Table~\ref{tab:downstream_demographics} presents the performance normalized scores for the Demographics downstream task group. We observe that BERT and DistilBERT have results close to each other in all tasks. In particular CoLES is very close in the age task but falls behind in the other two. Table~\ref{tab:downstream_risk} shows varied performance across Risk tasks. BERT excels in nsf and ret but struggles on rare events like stop and suf. DistilBERT is more balanced, with strong scores on stop and frozen, but weaker on unauth and suf. CoLES performs poorly overall, while FeatEng outperforms on some rare tasks. These results highlight the difficulty of modeling imbalanced events and the potential value of language model approaches. Table~\ref{tab:downstream_banking} shows that BERT and DistilBERT perform strongly on most classification tasks (debit card, fi, act type, act prof.), but both fail on inc. and bal., suggesting difficulty with high-cardinality predictions. CoLES excels in inc. and does moderately in bal., but underperforms elsewhere. FeatEng shows strong performance only on bal., with zero scores on all other tasks. Overall, transformer models dominate categorical tasks, while CoLES and FeatEng occasionally outperform on specific numeric or structured targets. Table~\ref{tab:downstream_geo} reports F1 scores for the Geolocation tasks. DistilBERT outperforms the other models across all tasks. BERT also performs well, with high scores across the board. CoLES performs poorly, and FeatEng fails completely. These results suggest transformer models are well-suited for geolocation inference, with DistilBERT slightly ahead of BERT.


\begin{table}[ht]
\centering
\begin{tabular}{lccc}
\toprule
  (Task) & \multicolumn{1}{c}{gender} & \multicolumn{1}{c}{1st\_name} & \multicolumn{1}{c}{age} \\
\midrule
 (Model) & acc & roc & acc \\
\midrule
BERT & 0.97 & 0.94 & 0.99 \\
DistilBERT & 1.00 & 1.00 & 1.00 \\
CoLES & 0.76 & 0.11 & 0.90 \\
FeatEng & 0.00 & 0.00 & 0.00 \\
\bottomrule
\end{tabular}
\caption{Demographics group downstream tasks normalized scores.}
\label{tab:downstream_demographics}
\end{table}

\begin{table}[ht]
\centering
\begin{tabular}{lcccccc}
\toprule
 (Task) & \multicolumn{1}{c}{nsf} & \multicolumn{1}{c}{stop} & \multicolumn{1}{c}{unauth} & \multicolumn{1}{c}{frozen} & \multicolumn{1}{c}{suf} & \multicolumn{1}{c}{ret} \\
\midrule
(Model) & roc & pr & pr & pr & pr & pr \\
\midrule
BERT & 1.00 & 0.00 & 0.71 & 0.47 & 0.37 & 1.00 \\
DistilBERT & 0.52 & 0.87 & 0.36 & 1.00 & 0.00 & 0.42 \\
CoLES & 0.35 & 0.56 & 0.00 & 0.54 & 0.26 & 0.03 \\
FeatEng & 0.00 & 1.00 & 1.00 & 0.00 & 1.00 & 0.00 \\
\bottomrule
\end{tabular}
\caption{Risk group downstream tasks normalized scores.}
\label{tab:downstream_risk}
\end{table}

\begin{table}[ht]
\centering
\begin{tabular}{lcccccc}
\toprule
 (Task) & \multicolumn{1}{c}{debit\_card} & \multicolumn{1}{c}{inc.} & \multicolumn{1}{c}{bal.} & \multicolumn{1}{c}{fi} & \multicolumn{1}{c}{act type} & \multicolumn{1}{c}{act prof.} \\
\midrule
(Model) & roc & roc & roc & f1 & f1 & f1 \\
\midrule
BERT & 1.00 & 0.05 & 0.04 & 1.00 & 0.97 & 1.00 \\
DistilBERT & 1.00 & 0.00 & 0.00 & 1.00 & 1.00 & 0.91 \\
CoLES & 0.35 & 1.00 & 0.40 & 0.11 & 0.58 & 0.17 \\
FeatEng & 0.00 & 0.06 & 1.00 & 0.00 & 0.00 & 0.00 \\
\bottomrule
\end{tabular}
\caption{Banking group downstream tasks normalized scores.}
\label{tab:downstream_banking}
\end{table}

\begin{table}[htb]
\centering
\begin{tabular}{lcccc}
\toprule
 & \multicolumn{1}{c}{state 1} & \multicolumn{1}{c}{city 1} & \multicolumn{1}{c}{state 2} & \multicolumn{1}{c}{city 2} \\
\midrule
run\_name & f1 & f1 & f1 & f1 \\
\midrule
BERT & 0.87 & 0.88 & 0.82 & 0.87 \\
DistilBERT & 1.00 & 1.00 & 1.00 & 1.00 \\
CoLES & 0.09 & 0.11 & 0.09 & 0.15 \\
FeatEng & 0.00 & 0.00 & 0.00 & 0.00 \\
\bottomrule
\end{tabular}
\caption{Geo. group downstream tasks normalized scores.}
\label{tab:downstream_geo}
\end{table}









\begin{acks}
 The authors of this work would like to thank Trustly for all support.
\end{acks}

\bibliographystyle{ACM-Reference-Format}
\bibliography{sample-base}



\end{document}